\documentclass[letterpaper]{article} 
\usepackage{aaai19}  
\usepackage{times}  
\usepackage{helvet}  
\usepackage{courier}  
\usepackage{url}  
\usepackage{graphicx}  
\usepackage{latexsym}

\usepackage{dcolumn}
\usepackage{latexsym}
\usepackage{amsmath}
\usepackage{amssymb}

\usepackage{multirow}
\usepackage[lined,boxed,commentsnumbered]{algorithm2e}
\usepackage{xcolor}

\usepackage{mathrsfs}

\usepackage{subfig}
\usepackage{amsfonts}
\usepackage{booktabs}
\usepackage{todonotes}

\usepackage{bm}

\usepackage{tikz}

\usepackage{mathtools}

\def\x{\mathbf{x}}

\def\bv{\mathbf{v}}
\def\bz{\mathbf{z}}

\def\h{\textbf{h}}
\def\e{\mathbf{e}}

\def\bb{\mathbf{b}}

\def\bW{\mathbf{W}}

\def\R{\mathbb{R}}

\def\x{\mathbf{x}}

\def\bv{\mathbf{v}}
\def\bz{\mathbf{z}}

\def\h{\textbf{h}}
\def\e{\mathbf{e}}

\def\bb{\mathbf{b}}
\def\Z{\mathbb{Z}}
\def\O{\mathbf{O}}

\def\H{\mathbf{H}}

\def\bW{\mathbf{W}}

\def\T{\mathcal{T}}
\DeclareMathOperator*{\argmax}{arg\,max}

\def\tran{^\mathrm{\scriptscriptstyle T}}

\def\o{\mathbf{o}}
\nocopyright

\title{Deformable Stacked Structure for Named Entity Recognition}
\author{Anonymous AAAI Submission}
\author{Shuyang Cao \and Xipeng Qiu \and Xuanjing Huang\\
Shanghai Key Laboratory of Intelligent Information Processing, Fudan University\\
School of Computer Science, Fudan University\\
825 Zhangheng Road, Shanghai, China\\
\{caosy14,xpqiu,xjhuang\}@fudan.edu.cn}

\begin{document}
\maketitle
\begin{abstract}
Neural architecture for named entity recognition has achieved great success in the field of natural language processing. Currently, the dominating architecture consists of a bi-directional recurrent neural network (RNN) as the encoder and a conditional random field (CRF) as the decoder. In this paper, we propose a deformable stacked structure for named entity recognition, in which the connections between two adjacent layers are dynamically established.
We evaluate the deformable stacked structure by adapting it to different layers. Our model achieves the state-of-the-art performances on the OntoNotes dataset.
\end{abstract}

\section{Introduction}
Named entity recognition(NER) is a subtask of sequence labeling. It is similar to other sequence labeling tasks considering its working procedure that is assigning
a certain label to each token of a sequence. But unlike part-of-speech(POS) tagging and other sequence labeling task evaluated on accuracy, the performance of
an NER system is evaluated on the whole named entity using precision, recall, and f1 score. Thus, the output of an NER system at each position is not independent
with each other and much related to its neighboring positions. The most common assumption of sequence labeling is the Markov property that the choice of label for a particular token is directly dependent only on the immediately adjacent labels; hence the labels of all the tokens in a sequence form a Markov chain.
Therefore, the widely-used statistical models for sequence labeling involve hidden Markov model (HMM) \cite{rabiner1986hmm},  maximum entropy Markov model (MEMM) \cite{DBLP:conf/icml/McCallumFP00} and conditional random field (CRF) \cite{lafferty2001conditional}.


In recent years, neural network architectures for NER \cite{huang2015bidirectional,P16-1101,D17-1283}
are proposed to reduce the efforts of feature engineering and the model complexity and have achieved great success. Currently, the dominative neural NER architecture consists of a bi-directional recurrent neural network (RNN) as the encoder and a conditional random field (CRF) as the decoder \cite{huang2015bidirectional}. The RNN encoder can effectively extract the context-aware features for each token, avoiding the cost of manually designing features.

Despite of their success, the network architecture of the encoder still need be manually designed for different tasks and lacks flexibility. Due to the inherent hierarchical structure of natural language, the crucial information for a token could appear in a changeable position. Taking the following NER instance for example, there are two entity mentions in the following sentence,
\begin{center}
  \textit{He bought 30 shares of \underline{Acme} in \underline{2006}}.
\end{center}%
For the token ``\textit{2006}'', its self-information is enough to decide its entity type. But for the token ``\textit{Acme}'', the information from its neighbour ``\textit{shares}'' could be more important. Since the position of the crucial information for each token is different, the current rigid network architecture heavily depends on the ability of RNNs to capture the context information.

In this paper, we propose a deformable stacked structure to flexibly choose the most informative features as input. Unlike the vanilla stacked structure, the connections between two adjacent layers are dynamically constructed. The input of each position in the upper layer is dynamically chosen from the lower layer, instead of a fixed position. Specifically, we introduce a dynamical offset to indicate the input's position. The offsets are dynamically computed according to the current hidden states. To make the whole neural network end-to-end trainable, we further propose an approximate solution to use a continuous offset to softly select the inputs via a bilinear interpolation, instead of the exact discrete offset. 


The contributions of the paper can be summarized as follows.
 \begin{enumerate}
   \item We propose a deformable stacked structure, whose stacking connections are dynamically determined, instead of in a pre-defined way. The deformable stacked way can effectively alleviate the pressure of RNNs for collecting the context information.
   \item We also propose an approximate strategy to softly change the connections, which makes the whole neural network differentiable and end-to-end trainable.
   \item  Compared to the models with rigid network architecture, our model is more flexible and suitable for named entity recognition tasks and achieves the state-of-the-art performances for named entity recognition on OntoNotes dataset.
 \end{enumerate}

\section{General Neural Architecture for Named Entity Recognition}

Given a sequence with $n$ tokens $X = \{x_1, \dots, x_n\}$, the aim of named entity recognition is to figure out the ground truth of labels $Y^* = \{y_1^*, \dots, y_n^*\}$:
\begin{equation}
Y^* = \argmax_{Y \in \mathcal{T}^n} p (Y | X), \label{eq:argmax}
\end{equation}
where $\mathcal{T}$ is the target set.

There are lots of prevalent methods to solve named entity recognition problem such as maximum entropy Markov model (MEMM), conditional random fields (CRF), etc.
Recently, neural models are widely applied to named entity recognition for their ability to minimize the effort in feature engineering
\cite{huang2015bidirectional,P16-1101}. Moreover, neural models also benefit from the distributed representations, which can enhance the generalization capabilities with the pre-trained word embeddings on the large-scale un-annotated corpus.

\begin{figure}[t]
 \centering
 \hspace{2em}
 \includegraphics[width=0.45\textwidth]{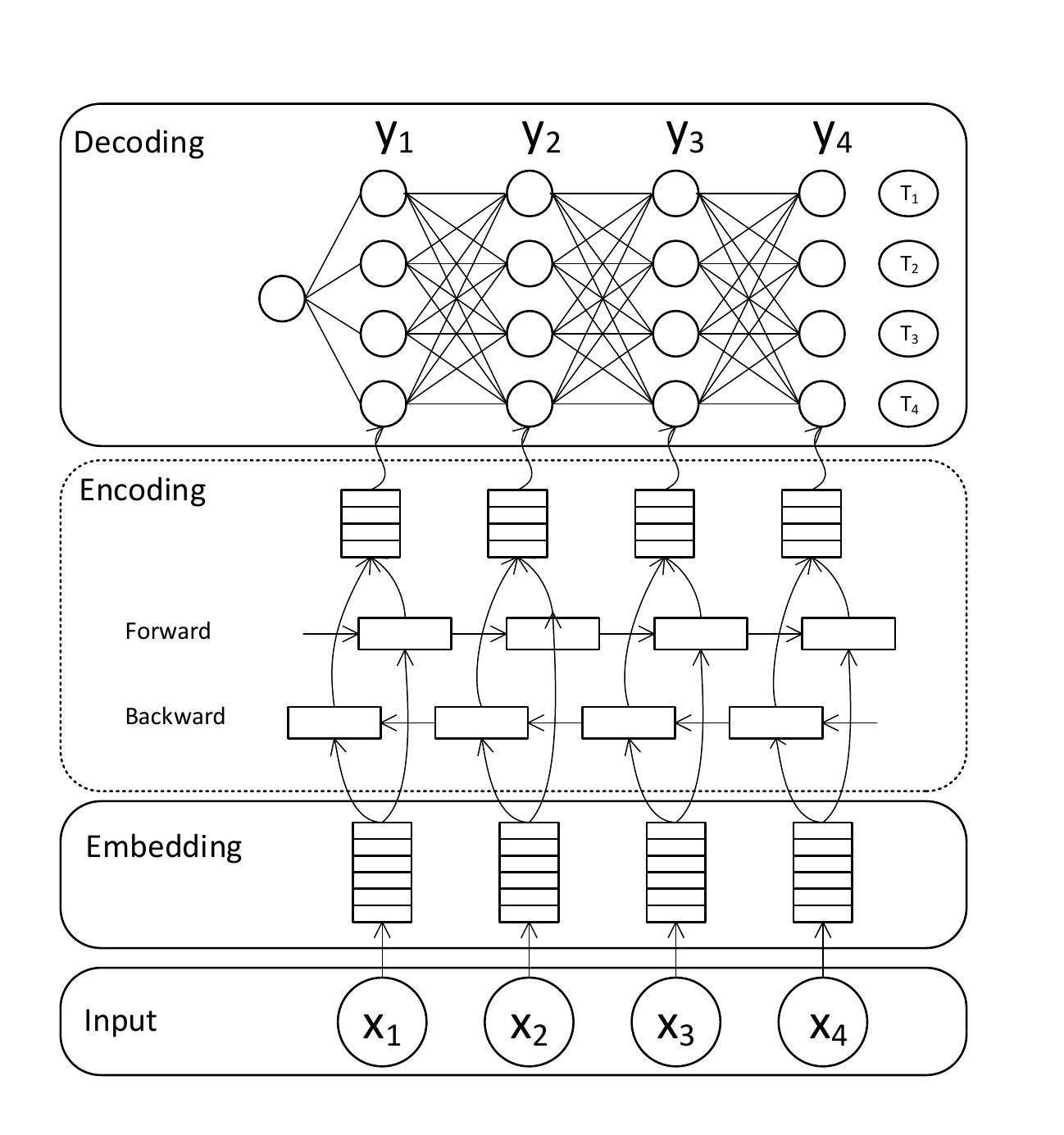}
 \caption{General neural architecture for named entity recognition. }\label{fig:arch}

\end{figure}

The general architecture of neural named entity recognition could be characterized by three components: (1) an embedding layer; (2) encoding layers consisting of several classical neural networks and (3) a decoding layer. The role of encoding layers is to extract features, which could be either convolution neural network or recurrent neural network. In this paper, we adopt the bidirectional long-short-term-memory (BiLSTM) neural networks followed by a CRF as decoding layer. Figure \ref{fig:arch} illustrates the general architecture.

\subsection{Embedding Layer}
To represent discrete tokens as distributed vectors, the first step is usually to map them to distributed embedding vectors. Formally, we lookup embedding vector from embedding matrix for each token $x_i$ as $\e_{x_i} \in \mathbb{R}^{d_e}$, where $d_e$ is a hyper-parameter indicating the size of embedding.

\subsection{Encoding Layers}

To incorporate information from both sides of sequence, we use bi-directional LSTM with forward and backward directions. 
Notably, the parameters of two LSTMs with different orientations are independent.
The update of each hidden state can be written precisely as follows:
\begin{align}
\h_i &= \overrightarrow{\h}_i \oplus {\overleftarrow{\h}_i},\\
&= \text{BiLSTM}(\e_{x_{1:n}}, i, \theta),
\end{align}
where $\overrightarrow{\h}_i$ and $\overleftarrow{\h}_i$ are the hidden states at position $i$ of the forward and backward LSTMs respectively; $\oplus$ is concatenation operation; $\theta$ denotes all the parameters in BiLSTM model.

\subsection{Decoding Layer}
After extracting features, we employ a conditional random fields (CRF)  layer to inference tags. In CRF layer, $p (Y | X)$ in Eq (\ref{eq:argmax}) could be formalized as:
\begin{equation}
p (Y | X) = \frac{\Psi (Y | X)}{\sum_{Y^\prime \in \T^n} \Psi (Y^\prime | X)}.
\end{equation}
Here, $\Psi (Y | X)$ is the potential function, and we only consider interactions between two successive labels (first order linear chain CRFs):
\begin{gather}
\Psi (Y | X) = \prod_{i = 2}^n \psi (X, i, y_{i-1}, y_i),\\
\psi (\x, i, y^\prime, y) = \exp(s(X, i)_{y} + b_{y^\prime y}),
\end{gather}
where $b_{y^\prime y} \in \R$ is transition parameter, indicating how possible a label $y^\prime$ will transfer to another label $y$. Score function $s(X, i) \in \mathbb{R}^{|\mathcal{T}|}$ assigns score for each label on tagging the $i$-th character:
\begin{equation}
s(X, i) = \bW_s^\top \mathbf{h}_i + \bb_s,
\end{equation}
where $\mathbf{h}_i$ is the hidden state of BiLSTM at position $i$; $\bW_s \in \mathbb{R}^{d_h \times |\mathcal{T}|}$ and $\bb_s \in \mathbb{R}^{|\mathcal{T}|}$ are trainable parameters.

At test phase, the Viterbi algorithm is employed to decode the best target sequence in polynomial time complexity.

\begin{figure}[t!]
  \centering
  \subfloat[Vanilla stacked structure]{
  \includegraphics[width=0.5\textwidth]{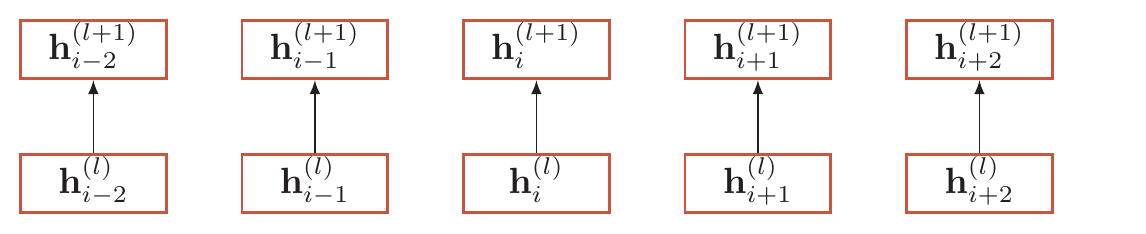} \label{fig:stacked-rnn}
  }
  \\
  \subfloat[Deformable stacked structure]{
  \includegraphics[width=0.5\textwidth]{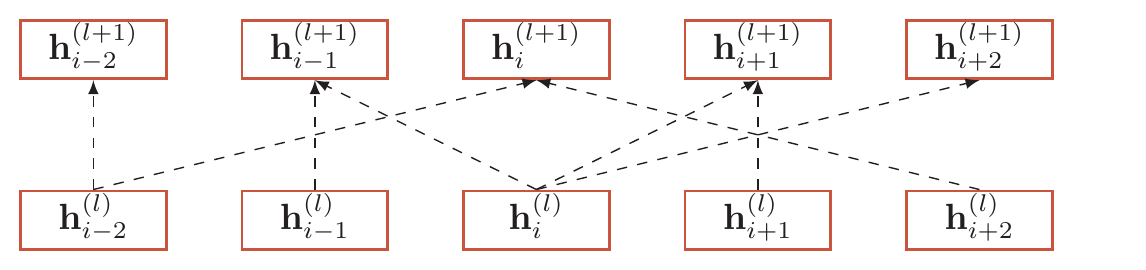} \label{fig:Model-III}
  }
  \caption{Two ways to stack layers. The dashed lines denote the connections are dynamically changed.}\label{fig:arch-comparison}
\end{figure}

\section{Deformable Stacked Structure}

The critical factor of neural named entity recognition models is the encoding layer, whose role is to extract useful features to judge the label of each token. To better model the complex compositional features, we could increase the depth of neural network by stacking the recurrent encoding layers.

\paragraph{Vanilla Stacked Structure}

A conventional way to stack layers is to take the output of the lower layer as the input of upper layer at each position \cite{pascanu2013construct}.
For each position $i$ at $(l+1)$-th layer, its input is taken from position $i$ at $l$-th layer.
\begin{align}
\h^{(l+1)}_i &= f(\h^{(l)}_i, \theta^{(l+1)})
\end{align}
where $f$ is a non-linear function.



Despite being successful for NER, the vanilla stacked structure has a limitation of the fixed geometric structures. The stacking structures are manually designed and lack flexibility.

\paragraph{Deformable Stacked Structure}

Due to the inherently hierarchical structure of natural language, the crucial information for a token could appear in a changeable position. As usually seen in NER, while in some case the information of a token itself is enough to decide its entity type, in another case the information of its neighbors could be more important.

To flexibly capture the most informative features, we propose a deformable stacked structure to choose the input's position from the lower layers dynamically.
For each position $i$ at $(l+1)$-th layer, its input is taken from position $i+o$ at $(l)$-th layer.
\begin{align}
\h^{(l+1)}_i &= f(\h^{(l)}_{i+o}, \theta^{(l+1)})
\end{align}
where $o$ is a offset, $o \in \Z$.

The offsets $o$ is predicted by an extra module based on the hidden states of the lower layers.

We can adapt deformable stacked structure between different layers, which we will further explain in the experiment section.

Figure \ref{fig:arch-comparison} gives the comparison of two different stacked structure.

\section{Differentiable Deformable Stacked Structure}

In our proposed deformable stacked structure, the index of the lower layer is discrete, which results in a non-differentiable network. Although we can use reinforcement learning to learn the parameters, in this paper, we propose a differentiable variant to utilize the strengths of back-prorogation of the neural network.

\begin{figure}[t]
 \centering
 \hspace{2em}
 \includegraphics[width=0.48\textwidth]{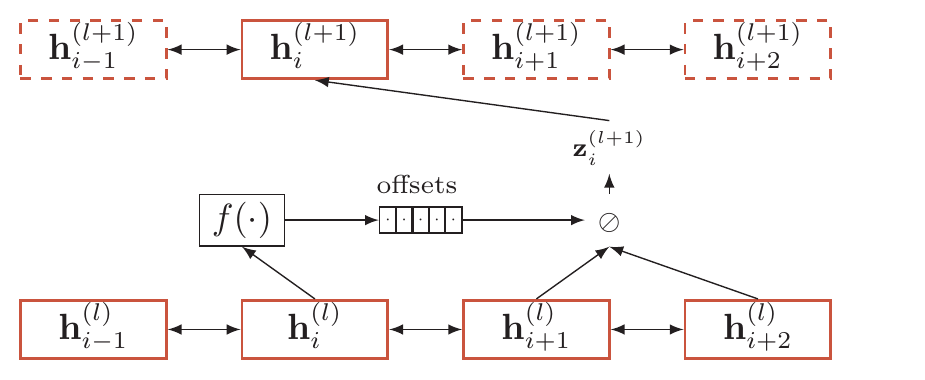}
 \caption{A differentiable implementation of deformable stacked structure. $f(\cdot)$ is defined by Eq (\ref{eq:offset}), $\oslash$ denotes the bilinear interpolation. }\label{fig:rnn-stacked-unfolded-deform-imp}

\end{figure}

Inspired by \cite{dai2017deformable}, we use bilinear interpolation to replace the exact discrete index.
Figure \ref{fig:rnn-stacked-unfolded-deform-imp} shows the
architecture of deformable stack.

\paragraph{Offset Learning}

Instead of a discrete offset, we use a continuous offset to make the whole network to be differentiable.
The offset is calculated by a simple function. The offset at the position $i$ can be given as:
\begin{equation}
  o_i^{(l)} = \bv \tran \h^{(l)}_i, \label{eq:offset}
\end{equation}
where $\bv \in \R ^{d}$ is the parameter vector, $\h^{(l)}_i \in \R^d$ is the hidden states at position $i$ of the $(l)$-th layer.

\paragraph{Deformable Input}

To bridge the continuous offset and discrete position, we apply bilinear interpolation to select the inputs from the lower layer softly.

For each position $i$ at $(l+1)$-th layer, its input $\bz^{(l+1)}_i$ can be given as:
\begin{equation}
  \bz^{(l+1)}_i = \sum_{j = 1}^{n} g ( i + o_i^{(l)} , j ) \cdot \h^{(l)}_j
\end{equation}
where $g( i + o_i^{(l)} , j )$ is a bilinear interpolation kernel and can be given as:
\begin{equation}
  g ( i + o_i^{(l)} , j ) = \max  ( 0, 1 -  |
  i + o_i^{(l)} - j |  ),
\end{equation}
thus $\{g(i + o_i^{(l)} , j)\}_{j=1}^n$ can be regarded as a mask vector which has two non-zero elements at most.

For example, assuming that the length of input sequence is 5, and the current position is 2, a continuous offset $o=1.2$ gives the mask vector
 \begin{align}
\mathbf{g} = [0,0,0.8,0.2,0].
\end{align}

By the above strategy, we can make the whole neural network to be differentiable, which can be end-to-end trained efficiently.

\subsection{Multi-Offset Extension}

To make the input of each recurrent layer more flexible, we can allow it to choose information from the multiple positions in the lower layer.

Therefore, a simple extension of our model is the multi-offset deformable structure, which allows the model to utilize the information at different positions jointly.

In multi-offset extension, we can predict $k$ offsets at position $i$,
\begin{equation}
  \o^{(l)}_i = V \h^{(l)}_i,
\end{equation}
where $\o^{(l)}_i \in \R^k$ is a vector consisting of $k$ offsets, and $V \in \R ^{k \times d}$ is the parameter matrix.

With $k$ offsets, we can obtain $k$ deformable inputs $\bz^{(l+1)}_{i1}, \bz^{(l+1)}_{i2}, \cdots ,\bz^{(l+1)}_{ik}$ via bilinear interpolation. Then we concatenate these inputs
to get the final input of position $i$ to the next layer.
\begin{equation}
  \bz^{(l+1)}_i = \bz^{(l+1)}_{i1} \oplus \bz^{(l+1)}_{i2} \oplus \cdots \oplus \bz^{(l+1)}_{ik},
\end{equation}
where $\oplus$ indicates the concatenation operation.

\subsection{Wide-Window Extension}

Another extension of our model is to involve neighbor hidden states when calculating the offsets.

In wide-window extension, we can predict $k$ offsets at position $i$ with window size $w$ by a convolutional neural network,

\begin{equation}
    \o^{(l)}_i = W \h^{(l)}_{i:i+d}
\end{equation}
\begin{equation}
  \O^{(l)} = \mathrm{Conv}(\H, W, w),
\end{equation}
where $\O \in \R ^{n \times k}$ is a matrix consisting of $k$ offsets in $n$ positions. $W \in \R ^{w \times k \times d}$ is the parameter matrix of the convolutional neural network.

\section{Training}
Given a trainset $(X^{(n)}, Y^{(n)})_{n = 1}^N$, the objective is to minimize the cross entropy loss $\mathcal{L}(\theta)$:
\begin{equation}
\mathcal{L}(\theta) = \frac{1}{N}\sum_n \log p(Y^{(n)}|X^{(n)}) + \lambda \| \theta \| ^2,
\end{equation}
where $\theta$ represents all the parameters, $\lambda$ represents the regularizer factor.

We use stochastic gradient descent with a momentum of 0.9.
The initial learning rate is set according to the dataset and task.
To avoid overfitting, dropout is applied after each recurrent or convolutional layer.

\paragraph{Initialization}
We take advantage of pre-trained word embeddings such as Glove \cite{pennington2014glove} to transfer more knowledge from large unlabeled data.
For the words that don't appear in Glove, we randomly initialize their embeddings from a normal distribution with mean 0 and standard deviation $\sqrt{\frac{1}{dim}}$ following \cite{P16-1101}, $dim$ is the dimension of word embedding.

The network weights are initialized with Xavier normalization \cite{glorot2010understanding} to maintain the variance of activations throughout the forward and backward passes. Biases are uniformly set to zero when the network is constructed.

\paragraph{Character Embedding}
Following \cite{P16-1101}, we also apply convolutional neural network (CNN) to extract character-level features of words.
The character-level feature of words helps the model better handle the OOV (out of vocabulary)
problems. For each character, we randomly initialize its embedding from a normal distribution
with mean 0 and standard deviation 1.

\section{Experiment}

We consider three different kinds of deformable stacked structure in our paper.
\begin{enumerate}
  \item Deformable stacked structure between BiLSTM layers.
  \item Deformable stacked structure between the encoder layer (BiLSTM in our paper) and decoder layer (CRF in our paper).
  \item Both 1 and 2.
\end{enumerate}


\begin{table}[t]
  \setlength{\tabcolsep}{2pt}
  \centering
  \begin{tabular}{lccc}
    \toprule
    \textbf{Dataset} & \textbf{Train} & \textbf{Dev} & \textbf{Test} \\
    \midrule
    {CoNLL-2003} & 204,567 & 51,578 & 46,666 \\
    \midrule
    {OntoNotes 5.0} & \multirow{2}*{1,088,503} & \multirow{2}*{147,724} & \multirow{2}*{152,728} \\
    {(CoNLL-2012)} \\
    \bottomrule
  \end{tabular}
  \caption{Number of tokens in different dataset.}
  \label{tab:dataset-detail}
\end{table}

\subsection{Datasets}

We evaluate our model on two datasets: CoNLL-2003 NER dataset \cite{tjongkimsang2003conll} dataset and
OntoNotes 5.0 \cite{W13-3516} dataset. We adapt structure 2 on CoNLL-2003 dataset and structure 1, 2, 3 on OntoNotes 5.0 dataset.

For both datasets, we perform the experiments on the English
portion.

For the OntoNotes 5.0 dataset, we split the
data according to the CoNLL-2012 shared task following \cite{Q16-1026}.
The Pivot Text portion is excluded because it lacks gold annotations for named entities.

The original tags are converted to tags of ``BIOES'' (begin, inside, outside, end, singleton) tagging.

The CoNLL-2003 dataset contains only 4 types of named entities: PERSON, LOCATION, ORGANIZATION, AND MISCELLANEOUS, while the OntoNotes 5.0
dataset contains 18 types of named entities, including works of art,
dates, cardinal numbers, languages, and events.
With BIOES tagging, we have 18 labels for
CoNLL-2003 dataset and 74 labels for OntoNotes 5.0 dataset(including padding label).

For the digits in both of the datasets, we replace them with digit 0.
The details of the two datasets are shown in Table \ref{tab:dataset-detail}.

\subsection{Hyper-parameters}
Hyper-parameters of our models are shown in Table \ref{tab:hyper-parameters-ner}.

For the word embedding, we use Glove pre-trained embedding of 100 dimensions on 6 billion words.

\subsection{Result}

\begin{table}[t!]
  \centering
  \setlength{\tabcolsep}{4pt}
  \begin{tabular}{lcc}
    \toprule
    & \textbf{CoNLL} & \textbf{OntoNotes} \\
    & \textbf{2003} & \textbf{5.0} \\
    \midrule
    Offset CNN window size & 3 & 3\\
    Number of offsets $k$ & 3 & 3\\
    Char-CNN filters & 30 & 30 \\
    Char-CNN window size & 3 & 3 \\
    Size of LSTM state  & 256 & 200 \\
    LSTM layers & 1 & 2 \\
    Learning rate & 0.008 & 0.005 \\
    Dropout & 0.5 & 0.5 \\
    Batch size & 10 & 8 \\
    \bottomrule
  \end{tabular}
  \caption{Hyper-parameters.}
  \label{tab:hyper-parameters-ner}
\end{table}

\begin{table}[t!]\setlength{\tabcolsep}{3pt}
  \centering
  \begin{tabular}{lccc}
    \toprule
    \textbf{Model} & \textbf{P} & \textbf{R} & \textbf{F} \\
    \midrule
    \cite{N09-1037}$^{\star}$ & 84.04 & 80.86 & 82.42 \\
    \cite{W09-1119} & 82.00 & 84.95 & 83.45 \\
    \cite{Q14-1037} & 85.22 & 82.89 & 84.04 \\
    \cite{chiu2016sequential} & 86.16 & 86.65 & 86.40 \\
    \cite{D17-1283} &  - & - & 86.84 \\
    \cite{D17-1283}$^{\star\star}$ &  - & - & 86.99 \\
    \midrule
    BiLSTM-CNN-CRF & 87.03 & 86.97 & 87.00\\
    \midrule
    Deformable stacked structure & \textbf{88.02}& \textbf{88.01} & \textbf{88.01} \\
    \bottomrule
  \end{tabular}
  \caption{Performances on OntoNotes 5.0 dataset. $^{\star}$ denotes the result from \cite{W13-3516}. $^{\star\star}$ denotes the baseline from their paper. \cite{N09-1037}: joint parsing
  and NER model. \cite{W09-1119}: using many resources, such as Wikipedia, non-local features. \cite{Q14-1037}: combining coreference resolution, entity linking, and NER into a
  single CRF model with cross-task interaction factors. \cite{chiu2016sequential}: BiLSTM-CRF network with with many composite features. \cite{D17-1283}: iterated dilated CNN and CRF.}
  \label{tab:compare-ner-ontonotes}
\end{table}

\begin{table}[th]\setlength{\tabcolsep}{3pt}
  \centering
  \begin{tabular}{lccc}
    \toprule
    \textbf{Model} & \textbf{P} & \textbf{R} & \textbf{F} \\
    \midrule
    \cite{collobert2011natural} & - &- &86.96\\
    \cite{D15-1104} & - & - & 91.20 \\
    \cite{N16-1030} & - &- & 90.33    \\
    \cite{P16-1101} & 91.35 & 91.06 & 91.21 \\
    \cite{D17-1283} &  - & - & 90.54 \\
    \midrule
    Bi-LSTM-CNN-CRF  &  {91.03} & {91.11} & {91.07} \\
    \midrule
    Deformable stacked structure & 91.01 & 91.24 & 91.12 \\
    \bottomrule
  \end{tabular}
  \caption{Performances on CoNLL-2003 dataset.
  \cite{collobert2011natural}: a earlier neural model. \cite{D15-1104}:
joint NER/entity linking model. \cite{N16-1030}: BiLSTM-CRF with RNN-based char information. \cite{P16-1101}: BiLSTM-CRF with CNN-based char information. 
\cite{D17-1283}: iterated dialated CNN  and CRF.}
  \label{tab:compare-ner-conll}
\end{table}

\begin{table*}[t]
  \centering
  \setlength{\tabcolsep}{4pt}
  \begin{tabular}{l|ccc|ccc}
    \toprule
    \multirow{2}*{\textbf{Model}} & \multicolumn{3}{c|}{\textbf{CoNLL-2003}} & \multicolumn{3}{c}{\textbf{OntoNotes 5.0}} \\
    & \textbf{P} & \textbf{R} & \textbf{F} & \textbf{P} & \textbf{R} & \textbf{F} \\
    \midrule
    BiLSTM-CRF & 87.97 & 86.49 & 87.22 & 85.26 & 85.87 & 85.56 \\
    \midrule
    Deformable stacked structure & 87.91 & 86.79 & 87.33 & 85.79 & 86.14 & 85.96\\
    \bottomrule
  \end{tabular}
  \caption{Performances of models without character embedding on test set.}
  \label{tab:compare-no-char}
\end{table*}

\begin{table}[t]
  \centering
  \setlength{\tabcolsep}{4pt}
  \begin{tabular}{ccccc}
    \toprule
    \textbf{Structure} & \textbf{\# of offsets} & \textbf{P} & \textbf{R} & \textbf{F} \\
    \midrule
    baseline & - &87.03 & 86.97 & 87.00 \\
    \midrule
    1 & 1 & 87.14 & 88.66 & 87.81 \\
    1 & 2 & 86.96 & 88.68 & 87.81 \\
    1 & 3 & 87.34 & 88.51 & 87.83 \\
    \midrule
    2 & 1 & 87.20 & 87.86 & 87.53 \\
    2 & 2 & 86.92 & 88.18 & 87.55 \\
    2 & 3 & 86.99 & 88.27 & 87.63 \\
    \midrule
    3 & 1 & 87.24 & 88.56 & 87.90 \\
    3 & 2 & 87.15 & \textbf{88.81} & 87.97 \\
    3 & 3 & \textbf{87.64} & 88.38 & \textbf{88.01} \\
    \bottomrule
  \end{tabular}
  \caption{Effects of different structure setting on OntoNotes 5.0 datset. Structure 1: deformable stacked structure between LSTM layers. Structure 2:
  deformable stacked structure between LSTM layers and CRF. Structure 3: both structure 1 and structure 2.}
  \label{tab:res-different-structure}
\end{table}

The results of our model on OntoNotes 5.0 and CoNLL-2003 dataset datasets are shown in Table \ref{tab:compare-ner-ontonotes} and Table \ref{tab:compare-ner-conll}.

On both of the two datasets, our model outperforms the vanilla stacked BiLSTM-CNN-CRF baseline. We also compare our model with existing models
on the two datasets. On OntoNotes dataset, our model also achieves the state-of-the-art result.

Deformable stacked structure utilizes the feature of neighbor words. Thus, it should make better predictions when OOV occurs.
We conduct experiments on models without character embedding. The results are shown in Table \ref{tab:compare-no-char}.




\begin{figure*}[t]
  \centering\small
    \subfloat[1 offset, structure 1]{
    \includegraphics[width=0.33\textwidth]{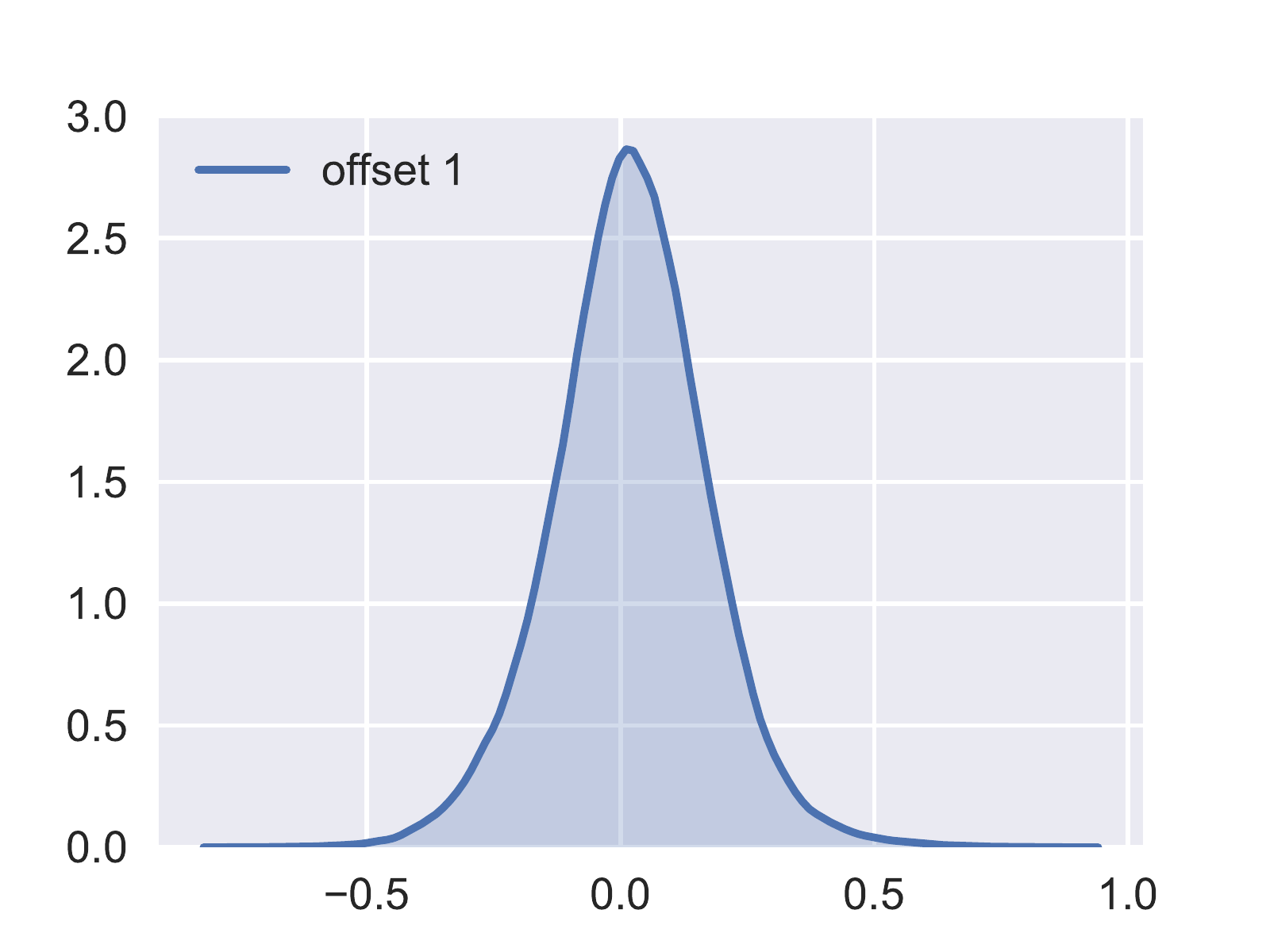} \label{fig:k1-bot}
    }
    \subfloat[2 offsets, structure 1]{
    \includegraphics[width=0.33\textwidth]{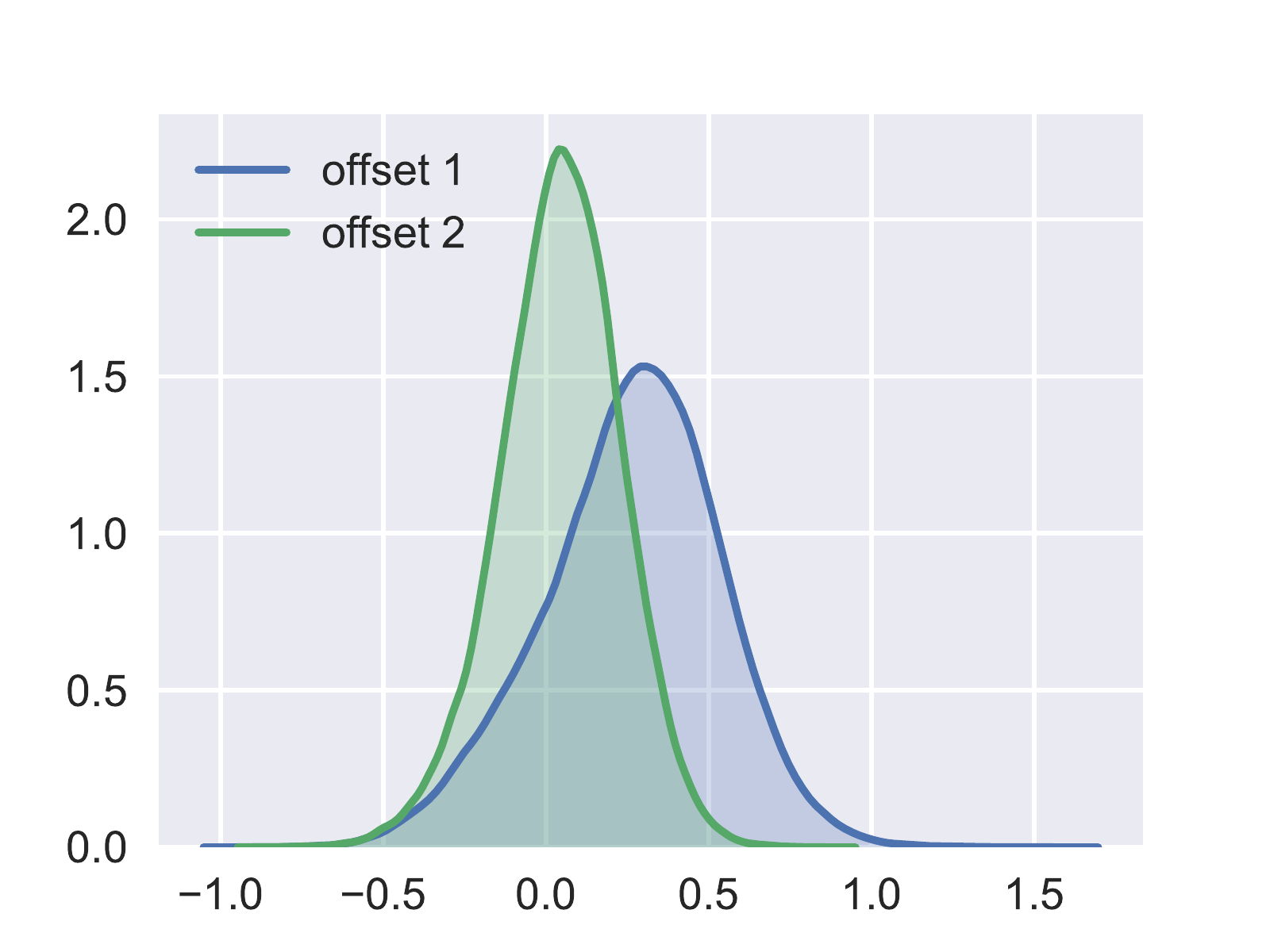}\label{fig:k2-bot}
    }
    \subfloat[3 offsets, structure 1]{
    \includegraphics[width=0.33\textwidth]{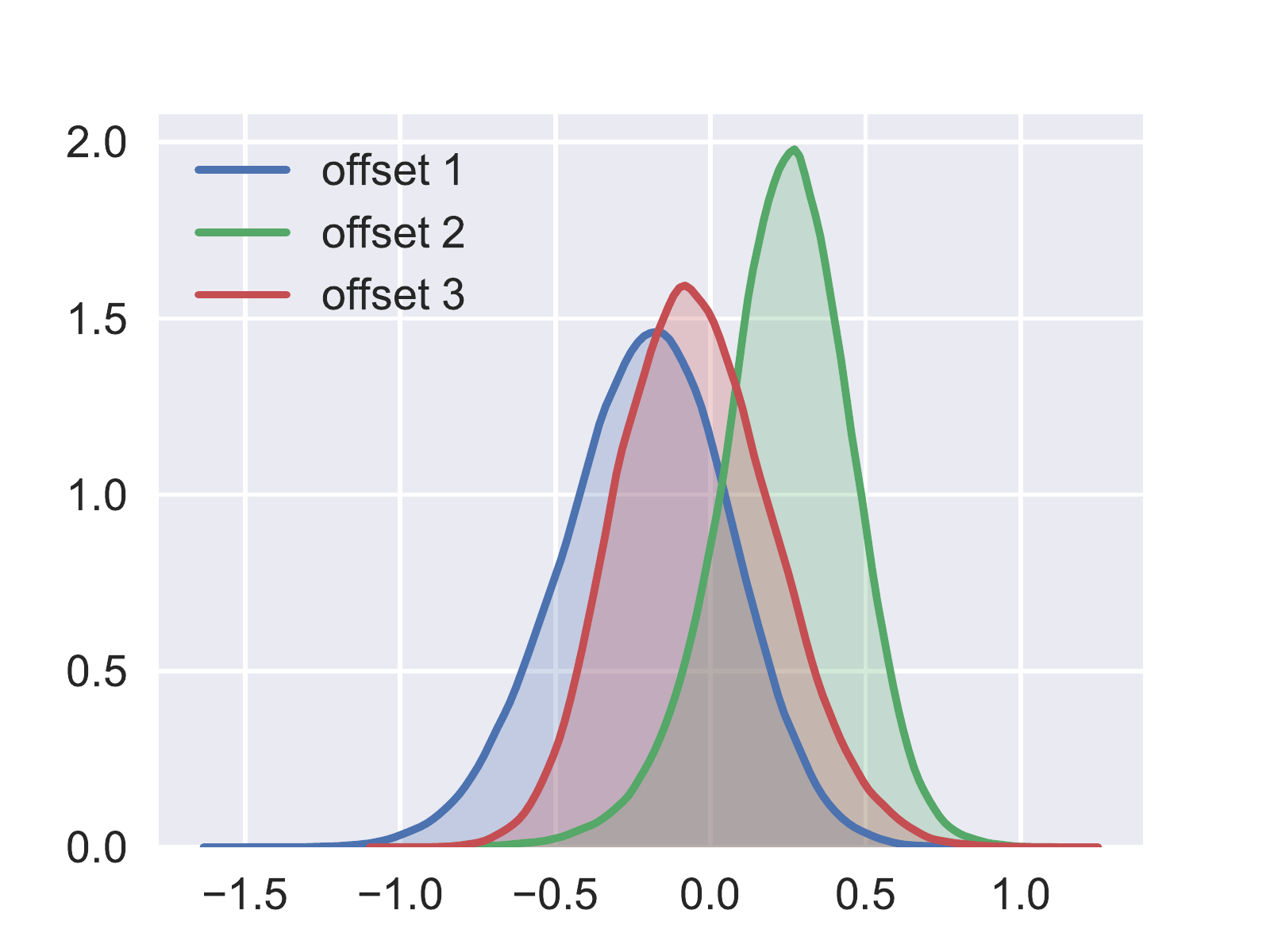} \label{fig:k3-bot}
    }
    \\\vspace{-1em}
    \subfloat[1 offset, structure 2]{
    \includegraphics[width=0.33\textwidth]{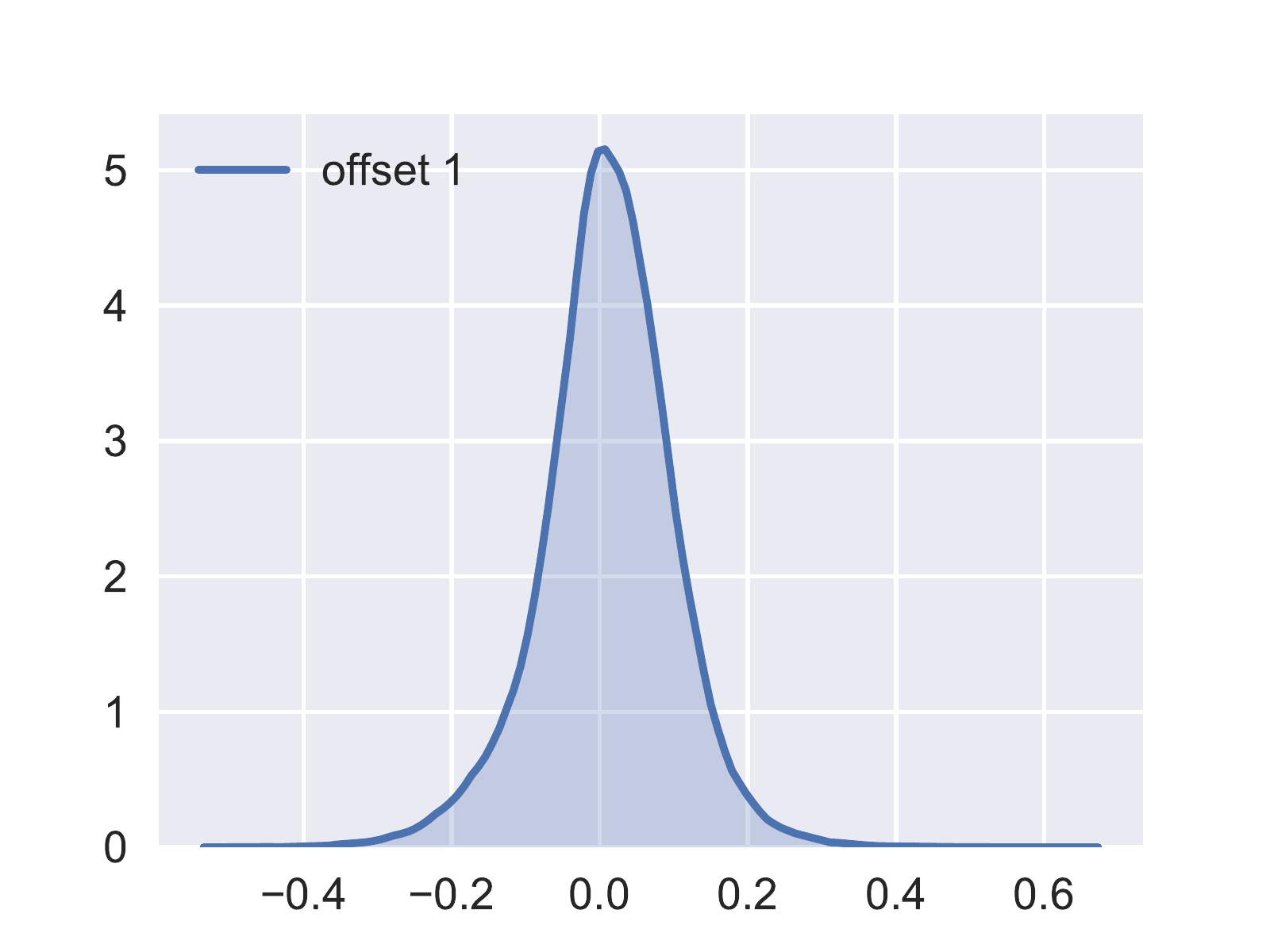} \label{fig:k1-up}
    }
    \subfloat[2 offsets, structure 2]{
    \includegraphics[width=0.33\textwidth]{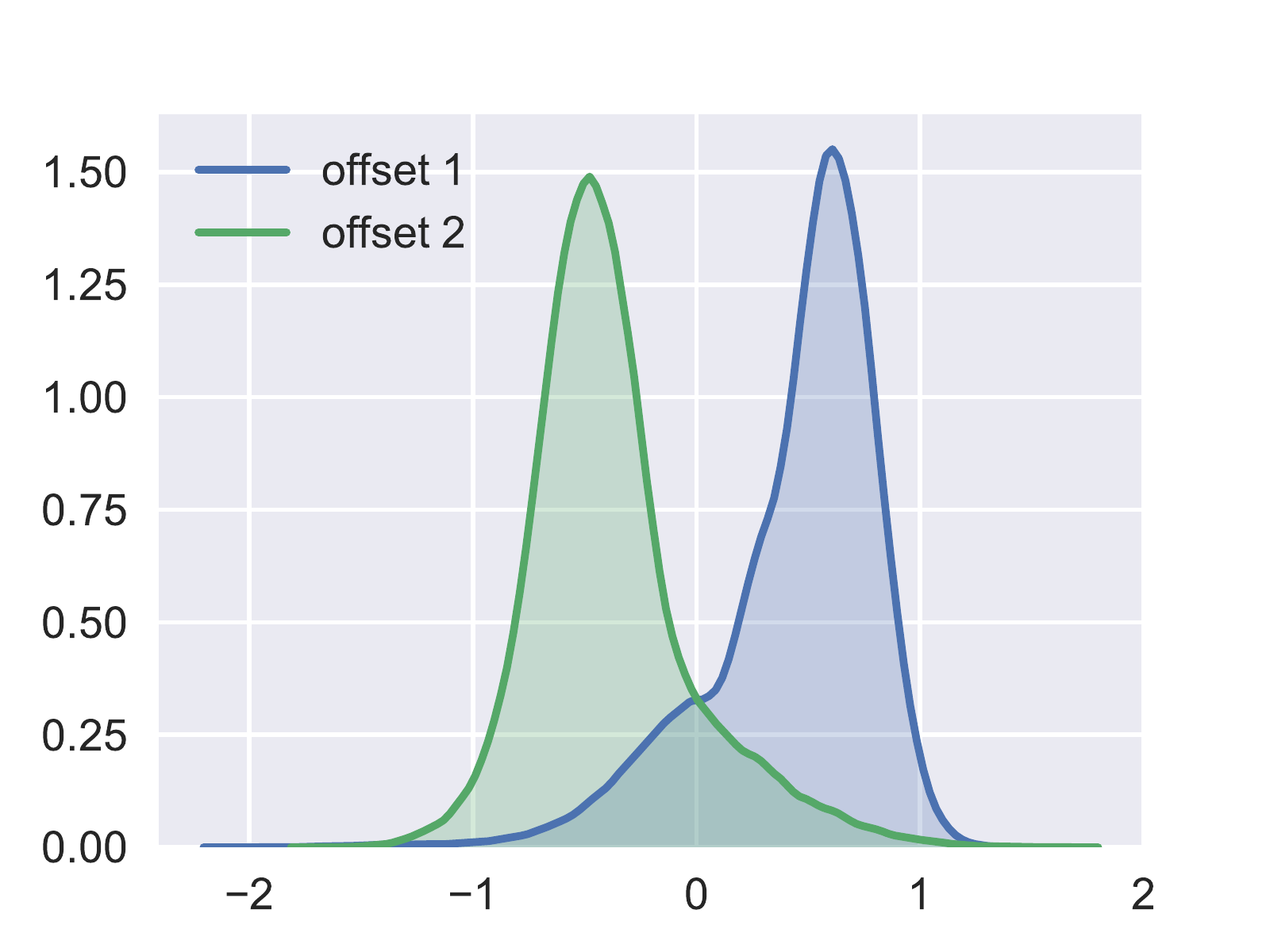} \label{fig:k2-up}
    }
    \subfloat[3 offsets, structure 2]{
    \includegraphics[width=0.33\textwidth]{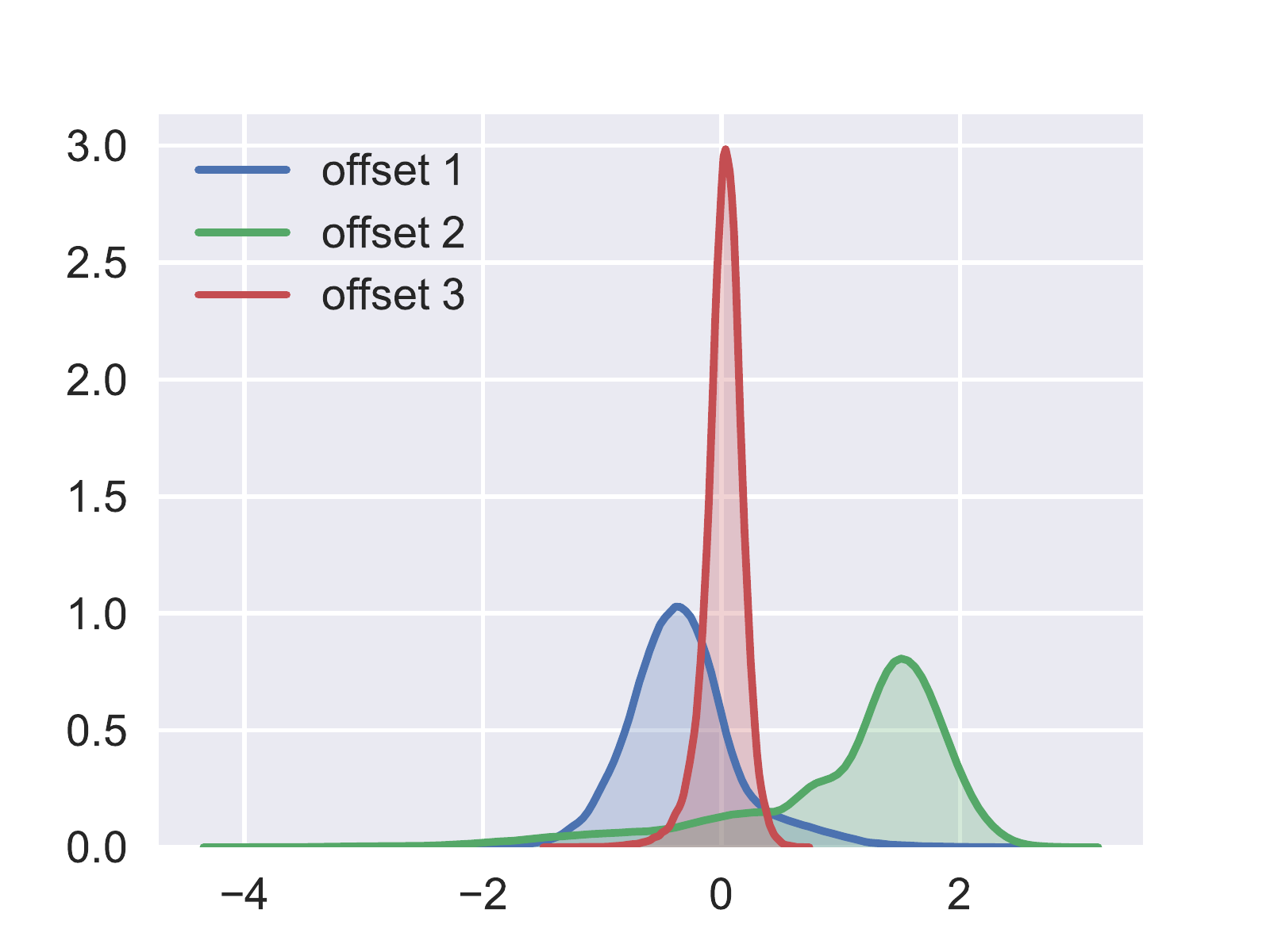} \label{fig:k3-up}
    }
    \caption{Kernel density estimation of offsets on test set. Y-axis represents the kernel density, and X-axis represents different values of offset.
    (a)(d): offset number $k=1$. (b)(e): offset number $k=2$. (c)(f): offset number $k=3$. Structure 1: deformable stacked structure between LSTM layers. Structure 2:
    deformable stacked structure between LSTM layers and CRF. Structure 3: both structure 1 and structure 2.}
    \label{fig:vis-offset}
  \end{figure*}

\paragraph{Comparison of different deformable stacked structure}

On OntoNotes dataset, we adopt three different deformable stacked structures. We evaluate the performance of the three different structure with varying
numbers of offsets.

As we can see from Table \ref{tab:res-different-structure}, the deformable stacked structure between LSTM layers has greater improvement
compared with the deformable stacked structure between the encoder layer and the decoder layer. These two structures differ on the function of their
next layers, and the difference in improvement also comes from that. For structure 1, the deformable stacked structure reconstructs the input of the next LSTM layer. Thus, it plays a role of feature augmentation
for the next LSTM layer to better extract related information. And better extraction of feature also helps with the decoding procedure. For structure 2, the deformable
stacked structure reconstructs the input of the CRF layer and plays a role of feature selection for decoding.

Taking information from multiple positions helps as increasing the number of offsets shows improvement.

We can also draw a significant character of the deformable stacked structure from Table \ref{tab:res-different-structure}. Compared with the baseline model,
the deformable stacked structure has great improvement in recall. It indicates that the deformable stacked structure can discover more named entity in the
sentence.

\subsection{Offsets Visualization}

To give an intuitive impression of how the offset dynamically changing, we show
the kernel density estimation of offset values of different deformable stacked structures on the OntoNotes test set in Figure \ref{fig:vis-offset}.

For the multiple offsets setting, we give the kernel density estimation of each offset. From Figure \ref{fig:vis-offset}, we can see that these offsets have the normal distribution in general.
For multiple offsets of deformable stacked structure between LSTM layers, we observe that at least one of the offsets has a distribution similar to
a normal distribution with mean 0 and the means of the rest distribution are slightly shifted from 0. For multiple offsets of deformable stacked structure
between the encoder layer and the decoder layer, the distributions shift from 0 significantly.

\subsection{Case Study}
\begin{figure*}[t]
    \centering
    \includegraphics[width=0.95\textwidth]{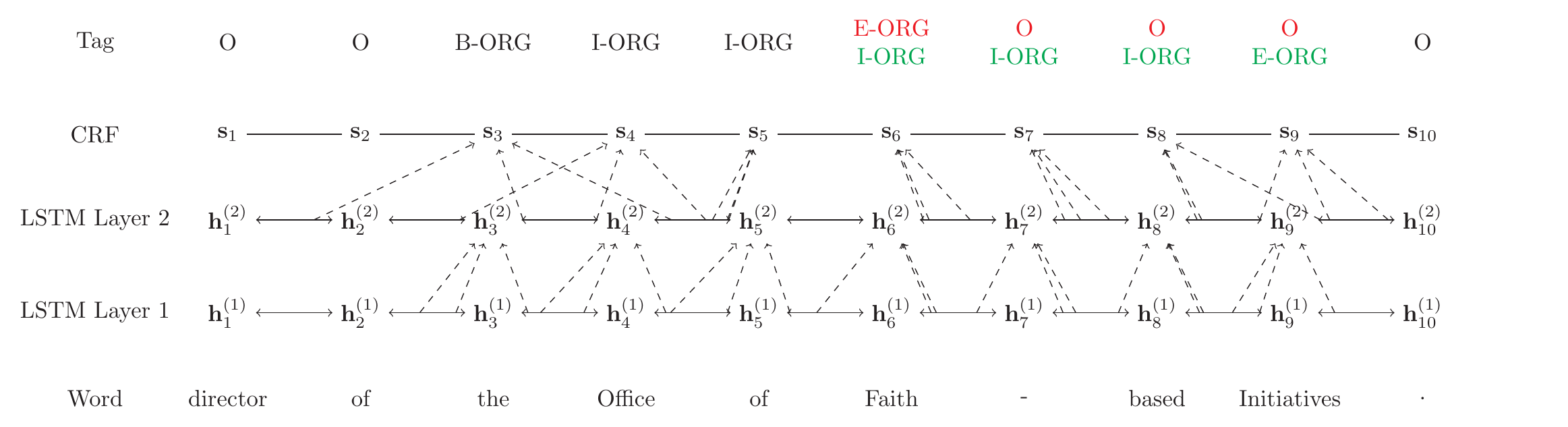}
    \caption{A real case on OntoNotes. The dashed lines indicate the dynamic offsets in our model. The red tags denote wrong predictions by the vanilla stacked structure, and the green tags denote the correct predictions by our
    model. The black tags denote the correct predictions by both models.}
    \label{fig:vis}
  \end{figure*}

We also visualize the offsets of a real case of structure 3 with offset number $k=3$ in Figure \ref{fig:vis}.
For simplicity, we only display the part of sentences that have different
predictions from the vanilla stacked model.


With deformable stacking, our model can better recognize the borders of named entities.
We can see that the positions that our model predicts correctly while the vanilla stacked structure predicts wrongly
have similar offsets between the CRF layer and the second LSTM layer. These offsets provide features from the
following positions and let the model know that the named entity doesn't break at the word "Faith".
Correctly predicting the borders of named entities shows great performance improvement when using BIOES tagging.

\section{Related Work}

There are mainly two lines of work related to ours.

One is the neural architecture for named entity recognition. Recently, several different neural network architectures have been proposed and successfully applied to NER. Among these neural architectures,
BiLSTM+CRF \cite{huang2015bidirectional} has become a fundamental architecture, which consists of a bi-directional LSTM as the encoder and a conditional random field (CRF) as the decoder. Some work \cite{Q16-1026,P16-1101,chen2017feature} also introduced a CNN layer before BiLSTM layer to model character-level information and achieved better performances. Besides BiLSTM, there is also some work to adopt CNN as encoder to capture the context information. \cite{D17-1283} use a dilated convolutional neural networks to efficiently aggregate broad context information.

Compared to these models, our model can effectively increase the input width of stacked layers and
help aggregate more broad context.

Another is neural architecture search \cite{pham2018efficient,zoph2016neural,AAAI1816537}.
Neural architecture search aims to automatically design the architecture of neural networks for a specific task. The current methods mainly adopt reinforcement learning to maximize the expected accuracy of the generated architectures on a validation set.

Our model can be regarded as a ``lightweight'' architecture search model, and changes the connections between the adjacent stacked layers. Moreover, we use an approximate strategy to change the connections softly.

\section{Conclusion}

We present deformable stacked structure, in which connections between two adjacent layers are dynamically generated. Three different deformable stacked structures
are designed and evaluated. Moreover, we also propose an approximate strategy to softly change the connections, which makes the whole neural network differentiable and end-to-end trainable.
Our model achieves the state-of-the-art performances on the OntoNotes dataset.

There are several potential directions for future
work. First, we hope to extend this work to build more flexible neural architecture. We have already established deformable connections between the most of layers of the encoder,
but some layers are still vanilla stacked such as the embedding layer.
Another exciting direction is to apply our model to other NLP tasks, such as parsing. Since our model does not require any task-specific knowledge, it might be effortless to apply it to these tasks.

\bibliographystyle{aaai}
\bibliography{nlp}
\end{document}